\documentclass[runningheads]{llncs}

\usepackage{eccv}

\usepackage{eccvabbrv}

\usepackage{graphicx}
\usepackage{booktabs}
\usepackage{algorithmic}
\usepackage{algorithm}
\usepackage{multirow}
\usepackage[table]{xcolor}
\usepackage{wrapfig}

\usepackage[accsupp]{axessibility}  

\usepackage{hyperref}

\usepackage{orcidlink}

\newcommand*{\affmark}[1][*]{\textsuperscript{#1}}

\begin{document}

\title{Through Van Gogh’s Eyes: Global Style Transfer with Diffusion Model} 

\titlerunning{Global Style Transfer}

\author{Jeongha Lee*\inst{1,2}\orcidlink{0009-0009-6932-2159} \and
Yujin Kim*\inst{3}\orcidlink{0000-0002-1442-3843} \and
Ghazanfar Ali\inst{4}\orcidlink{0000-0002-7741-1938} \and
Suhyun Kim$^{\dagger}$\inst{5}\orcidlink{0000-0003-0024-1704} \and 
Jae-In Hwang$^{\dagger}$\inst{1}\orcidlink{0000-0002-8748-9077} }

\authorrunning{J.~Lee et al.}


\institute{\affmark[1]Korea Institute of Science and Technology, \affmark[2]University of Science and Technology, \\ \affmark[3] Korea University, \affmark[4] Gachon University, \affmark[5] Kyung Hee University 
\\
\email{wjdgk1029@gmail.com, lakeeye1220@gmail.com, ghazan@gachon.ac.kr, dr.suhyun.kim@gmail.com, hji@kist.re.kr}
}

\onecolumn{%
\maketitle
\begin{center}
    \centering
    \captionsetup{type=figure}
    \includegraphics[width=1.0\linewidth]{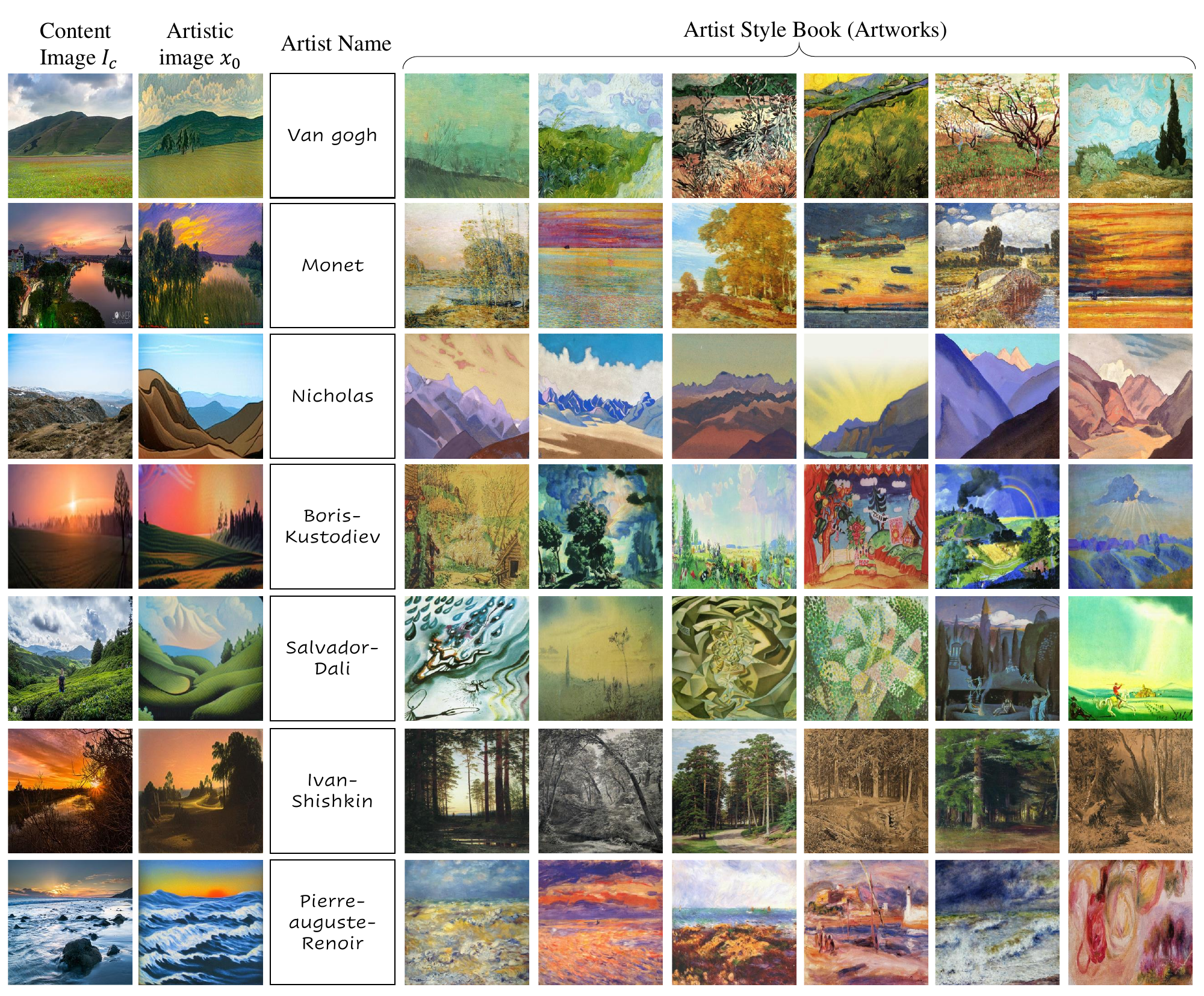}
    \captionof{figure}{Artist-level Global Style Transfer results. Given a content image $I_c$(first column), our framework synthesizes an artistic image
$x_0$ (second column) that reflects the global style of the target artist. The Artist Style Book (remaining columns) displays the top-6 real artworks from the artist's corpus with the highest semantic similarity to $x_{0}$, measured by CLIP distance. This demonstrates that our generated results are influenced by a diverse range of the artist's genuine works, successfully capturing the global stylistic distribution rather than overfitting to a single iconic exemplar.} 
\label{graphic_ab}
\end{center}%
}

\begin{NoHyper}
{
\renewcommand{\thefootnote}{\relax}
\footnotetext{$^*$ Equal contribution, $^\dagger$ Corresponding author}
}
\end{NoHyper}


\begin{abstract}

Artistic image synthesis aims to recreate the expressive visual identity of a target artist, yet existing methods often fail to capture an artist’s global style. Conventional style transfer methods transfer the style of one or a few reference artworks to a content image in a One-to-One manner, making them effective for artwork-level stylization but limited in representing the broader stylistic distribution of an artist. Text-to-image diffusion models conditioned on artist names, such as `\texttt{$\sim$ in Van Gogh style}', offer greater flexibility, but they often suffer from text-induced bias and reproduce patterns from only a few iconic works. To address these limitations, we introduce \emph{Global Style Transfer} (GST), an artistic image synthesis paradigm, in a Many-to-One manner,that aggregates multiple artworks from a target artist and transfers their shared global style to a single content image. For GST, we propose \emph{Global Style Guidance} (GSG), which learns a residual global style offset $\Delta \mathcal{h}_t$ in the intermediate feature space, or h-space, of a diffusion model under a fixed prompt. By learning artist-level style semantics purely from visual statistics, GSG mitigates text-dependent artistic bias. We further propose \emph{Content Alignment Guidance} (CAG), a training-free perceptual guidance mechanism that preserves the semantic structure of the content image while allowing artist-specific geometric deformation. Experiments on WikiArt demonstrate that GST achieves superior stylistic fidelity, content preservation, and output diversity compared to existing style transfer and diffusion-based artistic synthesis methods.

  \keywords{Global Style Transfer \and Diffusion Models \and Artistic Image Synthesis}
\end{abstract}

\section{Introduction}
\label{sec:intro}


If the great painters of the past were alive today, how would they perceive and depict our modern world on canvas? This question lies at the core of artistic image synthesis, which aims to recreate the expressive visual identity of a target artist. Existing approaches mainly fall into two paradigms: Style Transfer \cite{gatys2016image,huang2017arbitrary,li2017universal} and Text-to-Image (T2I) diffusion-based artistic image synthesis \cite{rombach2022high, zhang2023adding}. Conventional style transfer methods extract style from a single or a few reference artworks and transfer it to a content image, effectively reproducing the characteristics of specific artworks, called the \emph{One-to-One} approach. Unfortunately, 
such instance-level style transfer fails to capture the broader stylistic distribution that defines an artist’s overall visual identity, shown in \cref{GST}. On the other hand, T2I diffusion models conditioned on artist names (e.g., `\texttt{$\sim$ in the style of Van Gogh}') offer a more flexible synthesis pipeline but suffer from text-dependent artistic bias. T2I diffusion model often reproduces textures and compositions from a handful of iconic works, collapsing the stylistic diversity of an artist into a narrow, mode-biased distribution. Consequently, neither paradigm reliably captures the coherent visual identity that defines an artist's global style.  

To overcome these limitations, we introduce \emph{Global Style Transfer} (GST), a new paradigm that reformulates artistic image synthesis as a Many-to-One approach. Instead of relying on a single reference artwork or a text prompt, GST aggregates multiple artworks from a target artist and transfers their shared global style to a single content image, as shown in \cref{graphic_ab}. This formulation naturally mitigates instance-level bias by capturing coherent style representations across the full breadth of an artist's artworks, and eliminates linguistic dependence by grounding stylistic guidance entirely in visual statistics. As a result, GST enables artist-level style transfer that faithfully reflects the coherent visual identity of the target artist while allowing diverse and expressive content re-rendering.




Concretely, we realize Global Style Transfer through two complementary components. First, we propose \emph{Global Style Guidance} (GSG), which operates in the intermediate \emph{h}-space of a diffusion U-Net bottleneck. GSG trains a lightweight Style Extraction Function (SEF) to predict a residual global style offset $\Delta \mathbf{h}_t$ from multiple artworks under a fixed, unified text condition (e.g., `\texttt{A painting}'), thereby learning purely visual style semantics free from text-induced variance. Second, we propose \emph{Content Alignment Guidance} (CAG), a training-free mechanism that performs DDIM inversion \cite{song2020denoising} on the content image and applies perceptual guidance at each diffusion timestep. CAG preserves the abstract structure and semantic composition of the content while allowing style-driven geometric deformations, reflecting the expressive intent characteristic of a target artist.

Extensive experiments on WikiArt dataset \cite{tan2018improved} demonstrate that our framework achieves superior stylistic fidelity and content preservation compared to existing style transfer and T2I diffusion-based artistic synthesis methods, while producing diverse outputs that better reflect the global stylistic spectrum of the target artist. Our main contributions are as follows:

\begin{figure}[t!]
\centering
\includegraphics[width=0.7\linewidth]{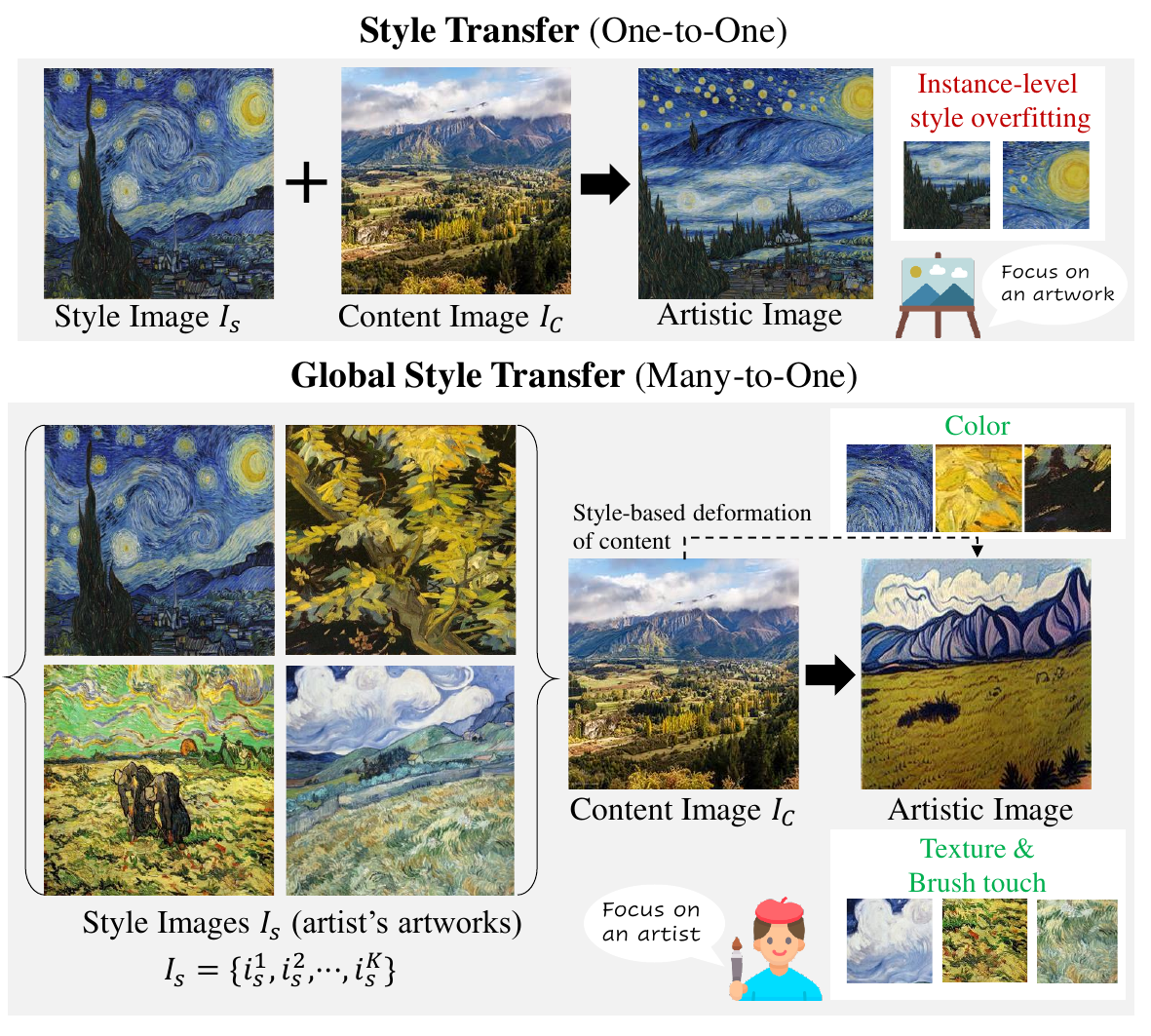}
\caption{Comparison between conventional style transfer and the proposed \emph{Global Style Transfer}.
(Top) Traditional style transfer applies the style of a single reference artwork to a content image, leading to instance-level overfitting. (Bottom) In contrast, \emph{Global Style Transfer} leverages various artworks from the same artist to learn unified stylistic features, including color palettes, texture patterns, and brushwork, enabling richer style-based content deformation and more faithful artistic image synthesis.}
\label{GST}
\end{figure}

\begin{itemize}

\item We introduce \emph{\textbf{Global Style Transfer}}, a novel many-to-one artistic image synthesis paradigm that learns a unified global style distribution from multiple artworks of a target artist, overcoming the artwork (instance)-level and text-dependent artistic bias of prior methods.

\item We propose \emph{\textbf{Global Style Guidance}} (GSG), a guidance mechanism operating in \emph{h}-space that learns artist-level global stylistic semantics only from visual statistics, mitigating linguistic dependence and artistic mode collapse.

\item We propose \emph{\textbf{Content Alignment Guidance}} (CAG), a training-free perceptual guidance that preserves the semantic integrity of the content while enabling artist-specific style-based deformation during the diffusion process.

\end{itemize}

\section{Related Works}
\label{sec:related_works}
\subsection{Style Transfer (1 content image, 1 style image)}
Style Transfer aims to generate a stylized image by combining the semantic structure of a content image with the visual characteristics (e.g., texture, color, and brushstroke) of a style image, typically in a \emph{One-to-One} setting (Content image: 1, Style image: 1) \cite{gu2018arbitrary,li2017universal}. Early Neural Style Transfer (NST) methods~\cite{gatys2016image,li2017demystifying} achieved this by matching feature statistics between content and style images, while later methods improved stylization quality through normalization \cite{ulyanov2016instance,huang2017arbitrary}, attention mechanisms \cite{park2019arbitrary,liu2021adaattn}. Recent diffusion-based approaches, such as StyleInjection \cite{chung2024style}, CSGO \cite{xing2024csgo}, InST \cite{zhang2023inversion} and Diff-NST \cite{ruta2024diff} further improve style fidelity by injecting style features into the denoising process. However, because these methods rely on only one or few style images, they capture instance-level style rather than the broader stylistic distribution of an artist.

\subsection{Style Personalization ($\times$ or 1 content image, Few style image)}
Recent style personalization methods focus on adapting pretrained text-to-image diffusion models to generate images in a customized artistic style, typically using a small number of reference styles and either no content image or a single content image to be stylized (Content: $\times$ or $1$, Style: few). These approaches fall into two categories. The first fine-tunes the diffusion backbone itself: 
DreamBooth \cite{ruiz2023dreambooth}, Custom Diffusion \cite{kumari2023multi}, and StyleDrop \cite{sohn2023styledrop} fine-tune the model, cross-attention projections, or lightweight adapters, respectively, to bind the style to a unique identifier. 
The second embeds style into the specific module without modifying model weights: Textual Inversion \cite{gal2022image} optimizes a word embedding for the target style, while StyleAligned \cite{hertz2024style} propagates stylistic consistency across generated images via shared attention at inference. However, these methods often memorize dominant visual patterns, limiting stylistic diversity.

In contrast to both methods, our \emph{Global Style Transfer} learns an artist-level global style distribution from multiple artworks and applies it to a content image in \emph{Many-to-One} manner, enabling faithful artistic synthesis without relying on single-instance style cues or text-dependent supervision.

\begin{figure*}[t!]
    \centering
    \includegraphics[width=1.0\linewidth]{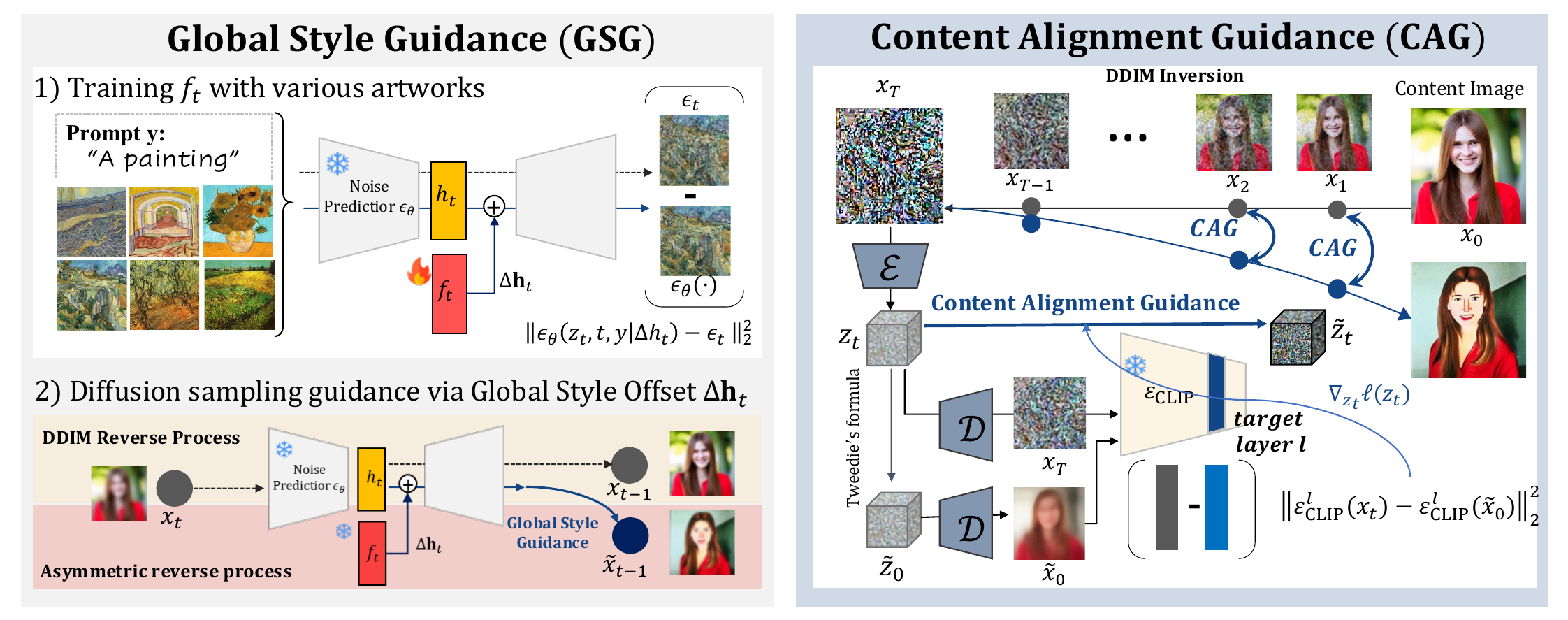}
    \caption{Overall framework of \emph{Global Style Transfer} (GST). (\textbf{Left}) \emph{Global Style Guidance} (GSG) trains a lightweight Style Extraction Function $f_t$ on multiple artworks under the fixed prompt `\texttt{A painting}', learning a residual global style offset $\Delta \mathbf{h}_t$ that modulates the U-Net bottleneck representation $h_t$ in a text-independent manner. During reverse diffusion, $\Delta \mathbf{h}_t$ steers the denoising trajectory toward the artist's global style distribution. (\textbf{Right}) \emph{Content Alignment Guidance} (CAG) first maps the content image $I_c$ into a noisy latent via DDIM inversion, then applies CLIP-based perceptual guidance at each timestep to align the semantic structure of the generated image with the content while permitting style-driven deformation, yielding artist-level stylization.} 
    \label{main}
\end{figure*}
\section{Methods}
\label{sec:method}
\subsection{Preliminary}
\subsubsection{Latent Diffusion Models.}
Latent Diffusion Models (LDM) \cite{rombach2022high} operate in a lower-dimensional latent space, enabling cost-efficient and scalable text-conditioned image synthesis. Given an image $x_0$, an autoencoder $\mathcal{E}: \mathbb{R}^k\rightarrow{\mathbb{R}^d}$ maps images into the latent vector $z_0=\mathcal{E}(x_0)$. A forward diffusion process gradually adds Gaussian noise according to a variance schedule $\beta_t$, yielding:
\begin{equation}\label{forward}
z_t=\sqrt{\bar{\alpha}_t}z_0+\sqrt{1-\bar{\alpha}_t}\epsilon, \ \epsilon\sim\mathcal{N}(0,I),
\end{equation}
where $\bar{\alpha}_t=\prod\nolimits_{s=1}^t(1-\beta_s)$. Then, the denoising model $\epsilon_\theta$ learns to predict the noise $\epsilon_\theta(z_t,t,c)$ given the noisy latent $z_t$ at every time step $t$ with text condition $c$ encoded by the text encoder $\tau_\phi$ during the sampling process as below:
\begin{equation}\label{DDIM}
    \mathbf{z}_{t-1} = 
    \sqrt{\bar{\alpha}_{t-1}}\tilde{z}_0
    + \sqrt{1-\bar{\alpha}_{t-1}-\sigma_t^2}\,\epsilon^t_\theta(z_t) 
    + \sigma_t \epsilon_t,
\end{equation}
where $\sigma_t=\eta\sqrt{(1-\bar{\alpha}_{t-1})/(1-\bar{\alpha}_t)}\sqrt{1-\bar{\alpha}_t/\bar{\alpha}_{t-1}}$. $\epsilon_\theta^t(z_t)$ denotes the model-predicted noise, abbreviated from $\epsilon_\theta(\mathbf{z}_t, t, c)$, and $\tilde{\mathbf{z}}_0$ means the approximation of the clean latent $\mathbf{z}_0$ at time step $t$ obtained via Tweedie’s formula. The training objective of the LDM is defined as:
\begin{equation}\label{SD}
\mathcal{L}_{\text{LDM}}:=\mathbb{E}_{z\sim\mathcal{E}(x),\epsilon_t,t}\left[ \lVert\epsilon_\theta(z_t,t,c)-\epsilon_t\rVert^2_2\right],
\end{equation}
After training, the diffusion process starts from a noisy latent $z_t$ and iteratively denoises it to $z_0$. The LDM then decodes $z_0$ into pixel space using the decoder $\mathcal{D}:\mathbb{R}^d\rightarrow\mathbb{R}^k$, yielding the clean image $x_0=\mathcal{D}(z_0)$.

\subsubsection{Asymmetric Reverse Process for Semantic Guidance.}
Asyrp \cite{kwon2022diffusion} discovers that the pre-trained diffusion model inherently contains a semantic space, referred to as the \emph{h}-space, corresponding to the activations of the U-Net bottleneck layer. To enable controllable generation within LDM, Asyrp reformulates the reverse process by modifying \cref{DDIM} into an asymmetric form as follows:
\begin{equation}\label{asymmetric}
    \mathbf{z}_{t-1} = 
    \sqrt{\bar{\alpha}_{t-1}}\mathbf{P}_t(\epsilon^t_\theta(z_t|\Delta \mathbf{h}_t))
    + \mathbf{D}_t(\epsilon^t_\theta(z_t))
    + \sigma_t \epsilon_t,
\end{equation}
where $\mathbf{P}_t$ denotes the predicted $\tilde{\mathbf{z}}_0$, and $\mathbf{D}_t$ represents the denoising direction toward $\mathbf{z}_t$. The asymmetric control updates only $\epsilon^t_\theta(\mathbf{z}_t)$ via $\epsilon^t_\theta(\mathbf{z}_t |\Delta\mathbf{h}_t)$, adding $\Delta\mathbf{h}_t$ as a residual to the U-Net bottleneck features $\mathbf{h}_t$, while preserving $\mathbf{D}_t$. This enables semantic manipulation in latent space without compromising reconstruction fidelity. Consequently, the \emph{h}-space exhibits stable and transferable semantics across time steps, supporting consistent, and multi-attribute guidance for latent diffusion models.
\begin{figure}[t!]
\centering
\includegraphics[width=0.8\linewidth]{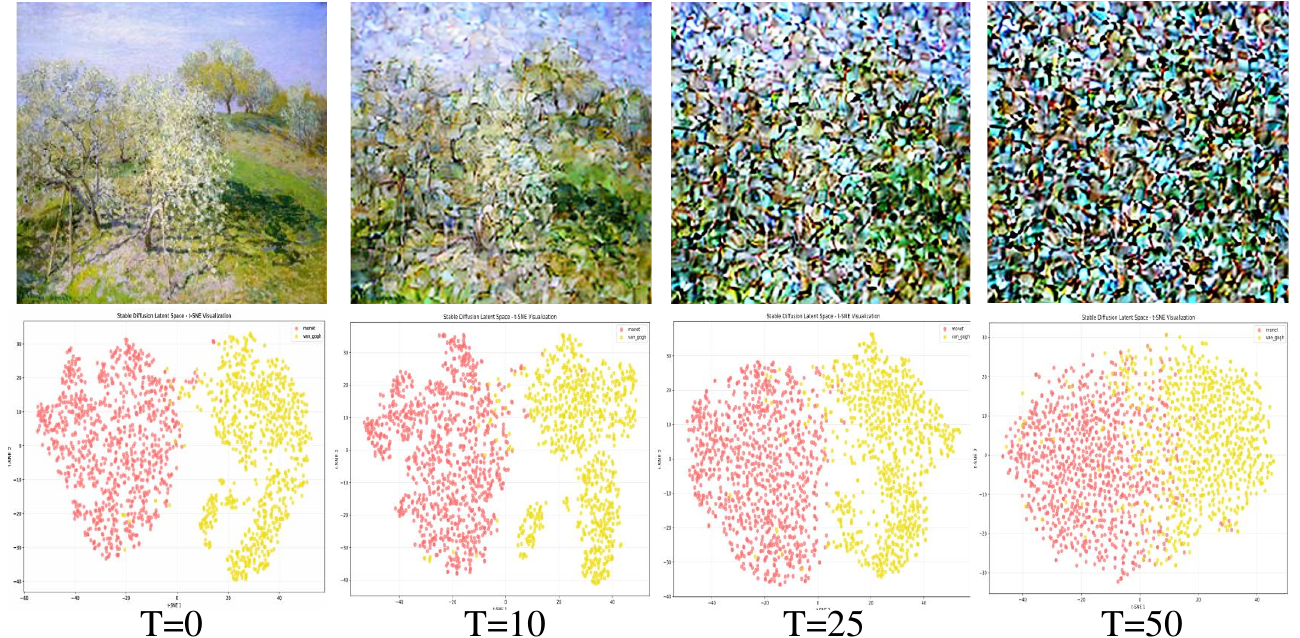}
\caption{Time-step robustness of \emph{Global Style Guidance} in \emph{h}-space. (\textbf{Top}) Artistic images generated at diffusion timesteps $t$ = \{0, 10, 25, 50\}, showing increasing noise levels as $t$ grows. (\textbf{Bottom}) t-SNE visualizations of 500 $h_t$ feature vectors from artworks by Van Gogh (yellow) and Monet (red) at the corresponding $t$. Despite progressive noise corruption, clusters remain well-separated across all timesteps, validating \emph{h}-space as a stable domain for artist-level style representation during the diffusion process.} 
\label{time_step}
\end{figure}
\subsection{Global Style Transfer}
We introduce \emph{Global Style Transfer}, a new paradigm that extends the conventional one-to-one style transfer setting to a many-to-one formulation, as illustrated in \cref{GST}. Unlike traditional methods that rely on a single style instance to guide the appearance of a content image, our approach aggregates multiple artworks from a single artist to learn a unified and consistent representation of that artist’s global style.

Given a content image $I_c$ and a set of style exemplars $I_s=\{i_s^1,i_s^2,\cdot \cdot, i_s^K\}$ drawn from an artist’s $K$ artworks, \emph{Global Style Guidance} (GSG) guides the diffusion model that encapsulates artist-specific semantic and style cues across various artworks. To preserve the semantic integrity of the content, we further introduce \emph{Content Alignment Guidance} (CAG), which maintains the abstract structure and semantics of $I_c$ while allowing flexible, style-based deformation. Through CAG, the content image is re-rendered with the distinctive visual characteristics of the target artworks without compromising its original composition.

\emph{Global Style Transfer} offers two key advantages over text-conditioned artistic synthesis.
(1) \textbf{Text-independent Guidance}: Rather than relying on textual prompts (e.g., `$\sim$ \texttt{in the style of Van Gogh}'), \emph{Global Style Transfer} extracts visual semantics directly from the aggregated style artworks. 
(2) \textbf{Artistic bias mitigation in diffusion-based artistic image synthesis}: Vanilla Text-to-Image diffusion models conditioned on artist names tend to exhibit stylistic bias, over-representing textures and compositions from a few iconic works. By guiding a global style prior from multiple style exemplars, our framework regularizes the diffusion process, mitigating such artistic mode bias and yielding diverse yet faithful images that better reflect the full stylistic spectrum of the artist.

\begin{algorithm}[!t]
\newcommand{\factorial}{\ensuremath{\mbox{\sc Factorial}}}
	\caption{Global Style Guidance Algorithm}\label{GSG_alg}
        \footnotesize
        {\bfseries Require:} Style exemplars (=artworks) $I_s$, Style Extraction Function (SEF) $f_t$, $\texttt{Model}$(Latent diffusion model) $\epsilon_\theta(\cdot, t)$, text encoder $\tau_\phi$, image decoder $\mathcal{D}$, Prompt $y$\\
        {\bfseries Output:} SEF $f_t$ 
        \begin{algorithmic}[1]
	\footnotesize
            \STATE $y$ =`\texttt{A painting}' 
            \COMMENT {All artworks are paired with this prompt}
            \FOR {$epoch=0$ \textbf{to} $E$}
            \FOR {$k$ = 1 \textbf{to} $K$}
            \STATE \textcolor{gray}{/*Diffusion Forward Process*/}
            \FOR {$t=0$ \textbf{to} $T-1$} 
            \STATE $z^k_0 =\mathcal{E}(I_s^k)$
            \COMMENT{instance style image sampling $i_s^k\in I_s$}
            \STATE $\epsilon_t\sim\mathcal{N}(0,I)$
            \STATE $z_t=\sqrt{\bar{\alpha}_t}z_0+\sqrt{1-\bar{\alpha}_t}\epsilon$
            \ENDFOR
            \STATE \textcolor{gray}{/*Diffusion Denoising Process*/}
		      \FOR {$t=T-1$ \textbf{to} $0$}
		        
                    \STATE $\mathcal{L}_{\text{SEF}}:=\mathbb{E}_{z\sim\mathcal{E}(x),\epsilon_t,t}\left[ \lVert\epsilon_\theta(z_t,t,\tau_\phi(y) |\Delta \textbf{h}_t)-\epsilon_t\rVert^2_2\right]$ \\
                    \STATE \textcolor{blue}{//$\Delta \mathbf{h}_t = f_t(h_t;\, \theta)$ from \cref{sef}}
                    \STATE $f_t(h_t;\, \theta) \leftarrow f_t(h_t;\, \theta) - \eta\nabla_{f_t}\mathcal{L}_{\text{SEF}}$  
                    \STATE $\mathbf{z}_{t-1} = 
                    \sqrt{\bar{\alpha}_{t-1}}\mathbf{P}_t(\epsilon^t_\theta(z_t|\Delta \mathbf{h}_t))
                    + \mathbf{D}_t(\epsilon^t_\theta(z_t))
                    + \sigma_t \epsilon_t$
            \ENDFOR \\
            \ENDFOR \\
            \ENDFOR \\
		\STATE \bfseries Return $f_t(h_t;\theta)$
    \end{algorithmic} 
\end{algorithm}

\begin{algorithm}[!t]
\newcommand{\factorial}{\ensuremath{\mbox{\sc Factorial}}}
	\caption{Global Style Transfer Algorithm}\label{GST_alg}
        \footnotesize
        {\bfseries Require:} Content Image $I_c$, $\texttt{Model}$(Latent diffusion model) $\epsilon_\theta(\cdot, t)$, text encoder $\tau_\phi$, image decoder $\mathcal{D}$, prompt $y$, CLIP image encoder $\mathcal{E}_{\text{CLIP}}$\\
        {\bfseries Output:} Artistic image $x_0$
        \begin{algorithmic}[1]
	\footnotesize
            \STATE $\mathbf{z}_T\leftarrow$ \text{DDIM Inversion} with respect to the $I_c$
		\FOR {$t=T-1$ \textbf{to} $1$}
		          \STATE $\epsilon_t\sim\mathcal{N}(0,I)$
                   \STATE \textcolor{blue}{//obtain $\Delta \textbf{h}_t$ from \cref{GSG_alg}}
                    \STATE $\mathbf{z}_{t-1} = 
                    \sqrt{\bar{\alpha}_{t-1}}\mathbf{P}_t(\epsilon^t_\theta(z_t|\Delta \mathbf{h}_t))
                    + \mathbf{D}_t(\epsilon^t_\theta(z_t))
                    + \sigma_t \epsilon_t$
                \STATE \textcolor{gray}{//Approximate $\tilde{z}_0$ from Tweedie formula \cite{efron2011tweedie} and $\tilde{x}_0\approx\mathcal{D}(\tilde{z}_0|z_t)$}
		        \STATE $\ell(z_t) = \|\mathcal{E}_{\text{CLIP}}^l(\tilde{x}_0) - \mathcal{E}_{\text{CLIP}}^l(x_t)\|_2$ 
                \STATE \textcolor{blue}{Content Alignment Guidance}
                    \STATE $\tilde{z}_t \leftarrow z_t - s\nabla_{z_t}\ell(z_t)$ 
                    \STATE \textcolor{gray}{//Classifier Free Guidance}
                    \STATE $\tilde{\epsilon}_t \gets \epsilon_\theta(\tilde{z}_t,t, \varnothing) + g (\epsilon_\theta(\tilde{z}_t,t, \tau_\phi(y)) - \epsilon_\theta(\tilde{z}_t,t, \varnothing))$ 
            \ENDFOR \\
		\STATE \bfseries Return $x_0=\mathcal{D}(\tilde{\mathbf{z}}_0)$
    \end{algorithmic} 
\end{algorithm}

\subsection{Global Style Guidance in \textit{h}-Space}
\subsubsection{Finding Artist's Global Style.}
To guide the diffusion model toward the global style of a specific artist, we define the residual global style offset $\Delta \mathbf{h}_t$, which modulates the U\textendash Net bottleneck representation $h_t$ at each timestep $t$:
\begin{equation}\label{hspace}
h_t = h_t + w \cdot \Delta \mathbf{h}_t,
\end{equation}
where $w$ controls the strength of stylistic modulation. The residual global style offset $\Delta \mathbf{h}_t$ captures the global stylistic semantics of the artist, enhancing the U-Net’s intermediate representations with consistent style attributes. To estimate $\Delta \mathbf{h}_t$, 
we define a \emph{Style Extraction Function} (SEF) $f_t$ that transforms the given $h_t$ into global style\textendash conditioned feature:
\begin{equation}\label{sef}
\Delta \mathbf{h}_t = f_t(h_t;\, \theta).
\end{equation}
$f_t$ is a lightweight multilayer perceptron (MLP) with one hidden layer parameterized by $\theta$ and conditioned on timestep $t$. Specifically, we apply the \emph{zero-initialization} strategy, which initializes the weights of $f_t$ to zero \cite{zhang2023adding} to avoid stochastic bias from random initialization. Zero-initialization ensures training process begins from an unbiased state and learns purely data-driven global style semantics. Then, we train the $f_t$ with multiple artworks using the noise reconstruction loss same as \cref{SD}:
\begin{equation}\label{SEF}
\mathcal{L}_{\text{SEF}}:=\mathbb{E}_{z\sim\mathcal{E}(x),\epsilon_t,t}\left[ \lVert\epsilon_\theta(z_t,t,\tau_\phi(y) |\Delta \textbf{h}_t)-\epsilon_t\rVert^2_2\right].
\end{equation}
Specifically, to ensure that training focuses purely on visual statistics rather than text conditioning, all artworks images $I_s$ are paired with the fixed simple prompt $y$, `\texttt{A painting}'. By enforcing a unified prompt across all artworks, the variance of text conditioning effectively approaches zero, allowing $f_t$ to rely purely on visual cues for capturing global style semantics.
Consequently, our SEF learns global stylistic representations directly from the visual semantics of the artworks rather than from linguistic information, resulting in visually grounded and unbiased \emph{Global Style Guidance}.
After training, the diffusion process proceeds toward a unified global style distribution, enabling the generation of images that reflect the coherent stylistic identity of the target artist without additional textual conditioning. The overall algorithms of GSG is described in \cref{GSG_alg}. 

\subsubsection{Time-Step Robustness of Global Style Guidance.}
To motivate our design choice of performing \emph{Global Style Guidance} (GSG) in the intermediate \emph{h}-space, we analyze whether \emph{h}-space provides a stable representation of global style across diffusion timesteps. As shown in \cref{time_step}, 500 artworks per artist form well-separated clusters in \emph{h}-space even under Gaussian noise across different timesteps. This demonstrates that $h_t$ remains robust to noise and timestep variation while preserving artist-level semantic global style, allowing GSG to maintain coherent stylization throughout the diffusion generative process.




\begin{table}[t!]
\caption{Comparison of implicit function learning between Asyrp and our framework.}
\centering
\resizebox{1.0\linewidth}{!}{
\begin{tabular}{lll}
\toprule
\textbf{Difference} & \textbf{Asyrp} & \textbf{Global Style Transfer (Ours)} \\
\midrule

Goal &
Instance-conditional attribute manipulation (Image Editing) &
Distribution-level global style training \\
\midrule
Supervision signal &
Text-driven attribute change &
Only visual statistics across artworks (Text-independent) \\
\midrule
\begin{tabular}[l]{@{}l@{}}
Text conditioning \\
on training $f_t$
\end{tabular}
&
\begin{tabular}[l]{@{}l@{}}
Explicit source--target prompts (high variance), \\
e.g., \texttt{`A girl'} ($y_\text{source}$) $\rightarrow$ \texttt{`A smiling girl'}($y_\text{ref}$)
\end{tabular}
&
\begin{tabular}[l]{@{}l@{}}
Fixed prompt (zero variance) \\
e.g., \texttt{`A painting'}
\end{tabular}
\\
\midrule
Training data &
Single (or few) source images &
Hundreds of artworks from a single artist \\
\midrule
\begin{tabular}[l]{@{}l@{}}
Loss function \\
for learning $f_t$
\end{tabular}
&
\begin{tabular}[l]{@{}l@{}}
CLIP directional editing loss + reconstruction loss: \\
$\displaystyle
\lambda_{\mathrm{CLIP}}
L_{\mathrm{direction}}
\left(
P_t^{\mathrm{edit}}, y_{\mathrm{ref}};\,
P_t^{\mathrm{source}}, y_{\mathrm{source}}
\right)
+
\lambda_{\mathrm{recon}}
\left\|
x_t^{\mathrm{edit}}
-
x_t^{\mathrm{source}}
\right\|_2$
\end{tabular}
&
\begin{tabular}[l]{@{}l@{}}
Noise reconstruction loss over artist distribution: \\
$\displaystyle
\mathbb{E}_{z \sim \mathcal{E}(x),\, \epsilon_t,\, t}
\left[
\left\|
\epsilon_\theta(z_t, t, \tau_\phi(y)\mid \Delta h_t)
-
\epsilon_t
\right\|_2^2
\right]$
\end{tabular}
\\

\bottomrule
\end{tabular}
}
\label{tab:asyrp}
\end{table}

\subsubsection{Difference between Asyrp and our Global Style Transfer.}
Although our framework shares the same implicit function architecture as Asyrp~\cite{kwon2022diffusion}, the two methods differ fundamentally in objective and supervision, as summarized in \cref{tab:asyrp}. Asyrp uses $f_t$ as an instance-level editing operator for single-image manipulation, optimized with a CLIP directional loss between source–reference text prompts (`\texttt{A girl}' $\rightarrow$ `\texttt{A smiling girl}'). Consequently, the optimization is inherently text-dependent, aligning $f_t$ with semantic attribute directions under highly varying prompt conditions. In contrast, we optimize $f_t$ to model the global stylistic distribution of an artist from hundreds of artworks using only visual supervision. By fixing the text condition $y$ to a constant prompt (e.g., `\texttt{A painting}'), we eliminate text-induced variance, allowing $f_t$ to be learned purely from artwork visual statistics via noise reconstruction.

\subsection{Content Alignment Guidance}
Motivated by how artists reinterpret real-world objects, preserving abstract structure while expressing unique artistic style, we propose the training-free \emph{Content Alignment Guidance} (CAG). In Neural Style Transfer (NST) \cite{ruta2024diff,zhang2022domain}, style-based deformation of content can be desirable, as artistic styles often involve intentional geometric or structural alterations beyond color and texture changes. Inspired by NST, our CAG aims to preserve the recognizable structure of the original content while allowing controlled, style-driven deformations that reflect the artist’s expressive intent.

We first perform DDIM inversion \cite{song2020denoising} to map the input content image into its noisy latent vector. 
Then, we measure perceptual similarity using the CLIP image encoder $\mathcal{E}_{\text{CLIP}}$ within a Stable Diffusion \cite{rombach2022high}  between the generated image $x_t=\mathcal{D}(z_t)$ at every timestep $t$ and approximated image $\tilde{x}_0\approx\mathcal{D}(\tilde{z}_0|z_t)$ via Tweedie’s formula \cite{efron2011tweedie}:
\begin{equation}
\ell(z_t) = \|\mathcal{E}_{\text{CLIP}}^l(\tilde{x}_0) - \mathcal{E}_{\text{CLIP}}^l(x_t)\|_2,
\end{equation}
where $\mathcal{E}_{\text{CLIP}}^l$ extracts high-level semantic features from layer $l$ of CLIP. Since the approximated image $\tilde{x}_0$ preserves the abstract structure of the content image, aligning its feature representation with that of the generated image $x_t$ encourages structural consistency while allowing style-driven deformation. Following the SDE formulation \cite{song2020score}, we incorporate this perceptual loss as the guidance by replacing the unconditional score $\nabla{z_t}\log p(z_t)$ with the conditional score $\nabla_{z_t}\log p(z_t|\ell)$, decomposed as:
\begin{equation}
\nabla_{z_t}\log p(z_t|\ell)
= \nabla_{z_t}\log p(z_t)
+ s\,\nabla_{z_t}\log p(\ell|z_t),
\end{equation}
where $s$ is a parameter controlling the guidance strength. The second term is interpreted as the gradient of the perceptual loss with respect to the image latent $\nabla_{z_t} \ell(z_t)$. Formally, the guided image latent is expressed as:
\begin{equation}
\tilde{z}_t = z_t - s\nabla_{z_t}\ell(z_t),
\end{equation}
where $s$ controls the guidance strength. Consequently, CAG refines the latent trajectory in a \emph{training-free} manner at each timestep to preserve content structure while enabling artist-specific compositional and geometric transformations. The overall algorithm and framework are illustrated in \cref{GST_alg,main}.

\section{Experiments}
\label{sec:experiments}
\subsubsection{Datasets.} 
We define the style reference images in the WikiArt dataset \cite{tan2018improved}, which includes over 80,000 artworks created by more than 1,000 artists across 27 art movements. 
For content images, we utilize the VanGogh2Photo dataset \cite{zhu2017unpaired} , which contains real-world photographic scenes paired with artistic counterparts, enabling consistent evaluation of content preservation in artistic synthesis. This dataset combination serves as a standard benchmark for the style transfer.
\subsubsection{Evaluation Metrics.} 
We evaluate using five metrics: FID~\cite{heusel2017gans} and ArtFID~\cite{wright2022artfid} for global style fidelity, CFSD~\cite{chung2024style} for content preservation, and CLIP-Div~\cite{alanov2023styledomain} and 1-Precision~\cite{kynkaanniemi2019improved, naeem2020reliable} for stylistic diversity and memorization, respectively. Detailed definitions are provided in the Appendix.

\begin{figure}[t!]
\centering
\includegraphics[width=1.0\linewidth]{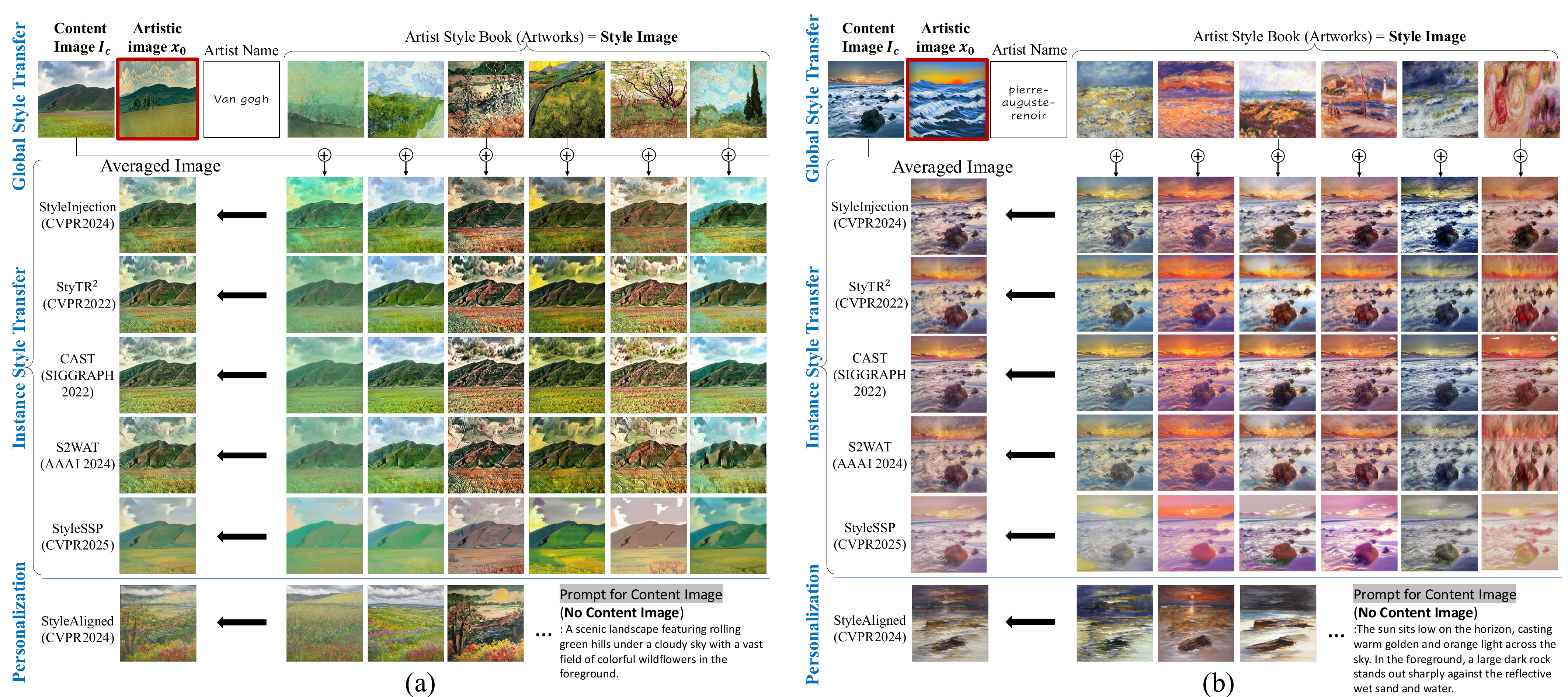}
\caption{Comparison of Global Style Transfer (GST) with representative style transfer baselines. We compare GST with instance-level style transfer and style personalization methods, using artworks from the same artist as style exemplars: (a) Van Gogh and (b) Pierre-Auguste Renoir. Since baseline methods utilize only a single style image $I_s$, we perform stylization separately using multiple artworks from the same artist and average the resulting outputs to approximate artist-level style. Baselines preserve only low-level appearance cues and produce inconsistent global styles, whereas GST learns the artist-level style and enables coherent global stylization.}
\label{style_transfer_baselines_fig}
\end{figure}

\subsection{Main Results}
\subsubsection{Sample Visualization.}
\cref{graphic_ab} shows the artistic images produced by the Global Style Transfer. This is achieved as the diffusion model performs \emph{Global Style Guidance} (GSG) within the intermediate feature space (\emph{h}-space), ensuring semantic stability across time steps and producing consistent style transformation throughout the generative process.

\subsubsection{Comparison with Style Transfer Methods.}
Our method introduces a new paradigm, \textit{Global Style Transfer} (GST), which fundamentally differs from conventional style transfer. Traditional style-transfer methods operate in a one-to-one setting, where a single content image $I_c$ is transferred to an artistic image  with only single style image $I_s$ through feature injection or attention modulation. While effective for instance-level stylization, such approaches cannot capture an artist-level style distribution spanning hundreds or thousands of artworks. To verify this limitation, we compare our method with representative Style Transfer baselines, including StyleInjection \cite{chung2024style}, StyTR$^2$ \cite{deng2022stytr2}, CAST \cite{zhang2022domain}, S2WAT \cite{zhang2024s2wat}, StyleSSP \cite{xu2025stylessp}, as well as the shared-attention personalization method StyleAligned \cite{hertz2024style}, grouped separately in \cref{style_transfer_baselines_fig}. Since these baselines take a single style image, we run each on the artist's artworks individually and average the resulting outputs to approximate an artist-level style.
As shown in \cref{style_transfer_baselines_fig}, this strategy fails to recover coherent artist-level style. Existing methods primarily capture low-level appearance cues such as dominant colors or coarse brush textures from each exemplar, but fail to preserve consistent stylistic characteristics shared across the artist’s full corpus. Even averaging multiple stylized outputs does not meaningfully represent the global style distribution, often producing blurred or inconsistent results. In contrast, our method directly learns the artist-level style manifold from multiple artworks, enabling coherent global stylization while avoiding overfitting to specific exemplars.

\begin{figure}[t!]
\centering
\begin{minipage}{0.5\linewidth}
    \captionof{table}{Quantitative comparison across artists. ArtFID measures style and content fidelity, CFSD measures content preservation, CLIP-Div and 1-Prec. assess stylistic diversity and memorization-avoidance. CLIP-Div is reported as \emph{Ours/Real} for direct comparison with the real artwork corpus. \textbf{Bold}: best; \underline{underline}: second best.}
    \label{compare}
    \resizebox{1.0\linewidth}{!}{%
    \begin{tabular}{clccccc}
    \toprule
    Artist & Method & FID ($\downarrow$) & ArtFID ($\downarrow$) & CFSD ($\downarrow$) & CLIP-Div ($\uparrow$) & $1-$Prec ($\uparrow$) \\ \midrule
    \multirow{5}{*}{Van Gogh} & S.D & 11.97 & 23.78 & 0.7428 & 0.178 & 0.257 \\
     & Textual Inversion & \textbf{7.67} & \textbf{13.68} & \underline{0.1674} & 0.220 & 0.550 \\
     & Custom Diffusion & 9.49 & \underline{14.71} & \textbf{0.1208} & \underline{0.289} & \underline{0.933} \\
     & LoRA based Fine-tuning & 11.01 & 21.38 & 0.1886 & 0.261 & 0.913 \\
     & \cellcolor{blue!10}{Ours} & \cellcolor{blue!10}{\underline{9.46}} & \cellcolor{blue!10}{19.25} & \cellcolor{blue!10}{0.2896} & \cellcolor{blue!10}{\textbf{0.297}/0.335} & \cellcolor{blue!10}{\textbf{0.988}} \\ \midrule
    \multirow{5}{*}{Chagall} & S.D & 16.26 & 32.44 & 0.3117 & 0.224 & 0.844 \\
     & Textual Inversion & \textbf{11.72} & \textbf{21.15} & 0.1683 & 0.238 & 0.877 \\
     & Custom Diffusion & 18.06 & 26.70 & \textbf{0.1108} & \underline{0.289} & \underline{0.962} \\
     & LoRA based Fine-tuning & 17.30 & 32.59 & \underline{0.1674} & 0.266 & 0.928 \\
     & \cellcolor{blue!10}{Ours} & \cellcolor{blue!10}{\underline{13.25}} & \cellcolor{blue!10}{\underline{26.39}} & \cellcolor{blue!10}{0.2626} & \cellcolor{blue!10}{\textbf{0.311}/0.293} & \cellcolor{blue!10}{\textbf{0.977}} \\ \midrule
    \multirow{5}{*}{Renoir} & S.D & 16.72 & 33.70 & 0.1584 & 0.225 & 0.987 \\
     & Textual Inversion & \textbf{12.14} & \textbf{21.99} & \textbf{0.1100} & 0.238 & 0.895 \\
     & Custom Diffusion & 15.15 & \underline{22.77} & \underline{0.1160} & \underline{0.294} & 0.987 \\
     & LoRA based Fine-tuning & 15.36 & 29.23 & 0.1749 & 0.279 & \underline{0.988} \\
     & \cellcolor{blue!10}{Ours} & \cellcolor{blue!10}{\underline{12.27}} & \cellcolor{blue!10}{24.70} & \cellcolor{blue!10}{0.1566} & \cellcolor{blue!10}{\textbf{0.318}/0.313} & \cellcolor{blue!10}{\textbf{0.999}} \\ \bottomrule
\end{tabular}}
\end{minipage}
\hfill
\begin{minipage}{0.48\linewidth}
    \centering
    \includegraphics[width=1.0\linewidth]{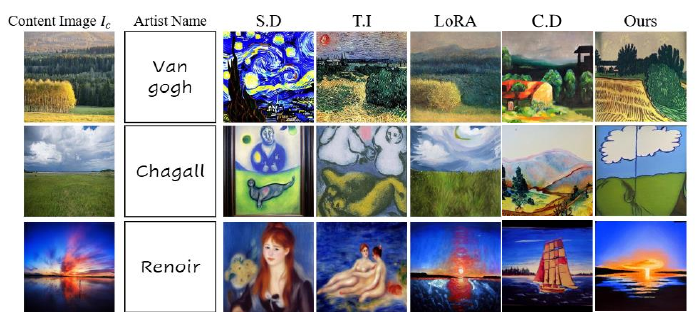}
\caption{
Visual comparison of previous style personalization methods and vanilla Stable Diffusion. Qualitative results corresponding to \cref{compare} are shown, demonstrating the comparative performance of each method in terms of style alignment and visual coherence.}
\label{compare_fig}
\end{minipage}
\end{figure}
\subsubsection{Comparison with Style Personalization Methods.}
We compare GST with representative personalization methods, including Textual Inversion (TI) \cite{gal2022image}, which optimizes a token embedding over the full artwork collection, DreamBooth \cite{ruiz2023dreambooth} with LoRA-based fine-tuning, and Custom Diffusion \cite{kumari2023multi}, which fine-tunes only the cross-attention layers. We also provide a qualitative comparison with StyleAligned~\cite{hertz2024style} in \cref{style_transfer_baselines_fig}. For a fair comparison in the GST setting, we train each baseline on the full artwork collection of each artist rather than using the few-shot setting adopted in their original formulations.  

As shown in \cref{compare}, GST achieves consistently strong performance across all three artists, demonstrating effective modeling of artist-level global style. GST achieves competitive ArtFID, indicating strong overall style and content fidelity without overfitting to content structure. It also achieves the highest CLIP-Diversity across all artists, showing that GST captures a broader stylistic distribution rather than collapsing to a narrow set of visual patterns. Notably, the CLIP-Diversity values of GST are highly consistent with those computed from the real artwork collections (reported as Ours / Real), suggesting that GST successfully reproduces the diversity of the true artist-level style distribution. The 1-Precision results further show that GST best avoids memorization while preserving stylistic consistency. In contrast, prior personalization methods often rely on memorized exemplar patterns or overfit to dominant visual motifs, as also observed in \cref{compare_fig}. For example, Textual Inversion is limited by its fixed-dimensional token embedding, while fine-tuning methods often collapse toward iconic compositions. Consequently, our method scales effectively with increasing artwork diversity and better captures the full artist-level style distribution. 


\begin{figure*}[t!]
\centering
\includegraphics[width=1.0\linewidth]{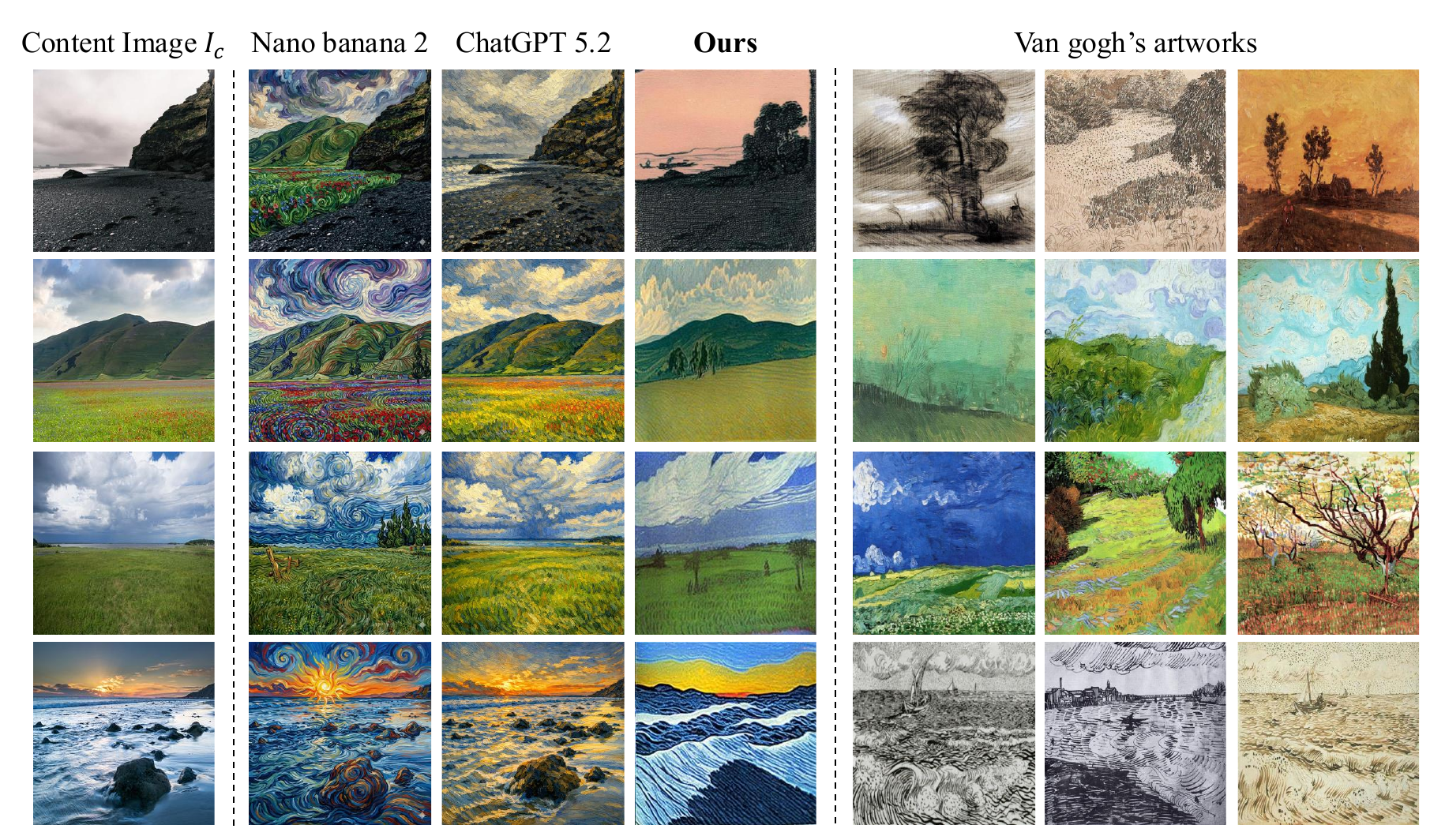}
\caption{Visual comparison with T2I models (Nano Banana 2, ChatGPT 5.2) prompted with `\texttt{Van Gogh style}' for a given content image $I_c$. The rightmost columns display the top three Van Gogh artworks with the lowest CLIP distance to our generated result.}
\label{vanilla}
\end{figure*}

\subsection{Analysis of Global Style Transfer}
\subsubsection{Analysis of Text-Independence in Global Style Transfer.}
We analyze whether the learned style representation depends on text prompts. 
\begin{wrapfigure}{r}{6cm}
\centering
  \includegraphics[width=1.0\linewidth]{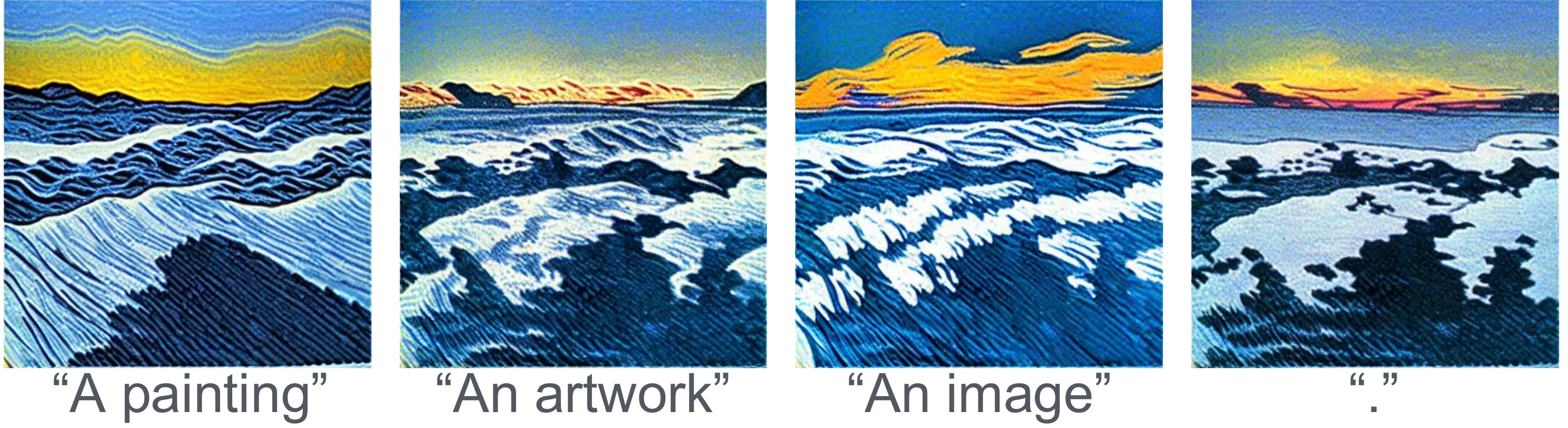}
  \caption{The sensitivity of prompt for training the Style Extraction Function $f_t$.}
\label{fig:prompt}
\end{wrapfigure}
Since the Style Extraction Function (SEF) is trained using a large collection of artworks, we use a fixed prompt, `\texttt{A painting}', during training to minimize textual influence and encourage the model to learn style primarily from visual cues. To evaluate prompt sensitivity, we replace the prompt with alternative generic prompts during training SEF and compute the corresponding style offset $\Delta h$. As shown in \cref{fig:prompt}, the generated images remain nearly identical across prompts, indicating that the learned representation is largely text-independent and primarily captures artist-level visual style.

\subsubsection{Analysis of Stylistic Bias in T2I Models.} 
We compare our framework with large-scale text-to-image baselines, Nano Banana 2 ~\cite{google_gemini_image_2024} and ChatGPT 5.2~\cite{openai_chatgpt_2024}, using the prompt `\texttt{Van Gogh style}'. As shown in \cref{vanilla}, these baselines exhibit strong stylistic bias, repeatedly generating images dominated by iconic Van Gogh features such as swirling patterns and thick brushstrokes, often resembling Starry Night. In contrast, our Global Style Transfer learns a unified artist-level representation, mitigating such bias and producing diverse outputs that better reflect the true global style distribution.

\begin{figure}[t!]
\begin{minipage}{0.49\linewidth}
    \centering
    \captionof{table}{Ablation study on Content Alignment Guidance (CAG). Removing CAG leads to a consistent increase in both FID and ArtFID, demonstrating its crucial role in preserving content fidelity under global style modulation}
    \resizebox{0.9\linewidth}{!}{%
    \begin{tabular}{lcc}
    \toprule
    \textbf{Method} & \textbf{FID} & \textbf{ArtFID} \\
    \midrule
    Ours w/o CAG & 11.05 & 22.45 \\
    \textbf{Ours} & \cellcolor{blue!8}\textbf{9.46} & \cellcolor{blue!8}\textbf{19.25} \\
    \bottomrule
    \end{tabular}}
    \label{tab:ablation}
\end{minipage}
\hfill
\begin{minipage}{0.49\linewidth}
    \centering
    \includegraphics[width=0.8\linewidth]{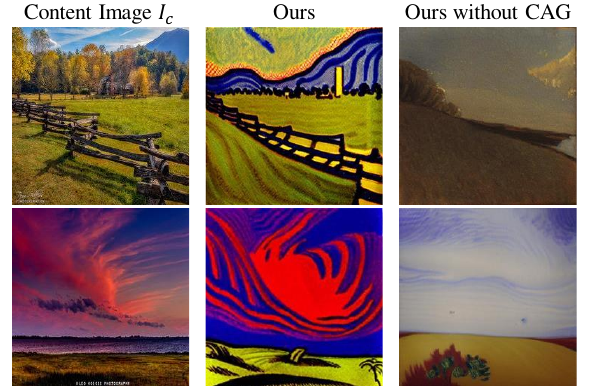}
    \captionof{figure}{Qualitative ablation on CAG. Without CAG, the images exhibit content degradation despite strong stylization.}
    \label{fig:ablation}
\end{minipage}

\end{figure}

\subsubsection{Effect of the Content Alignment Guidance.}
\begin{figure*}[t!]
\centering
\begin{minipage}{0.45\linewidth}
\centering
\includegraphics[width=1.0\linewidth]{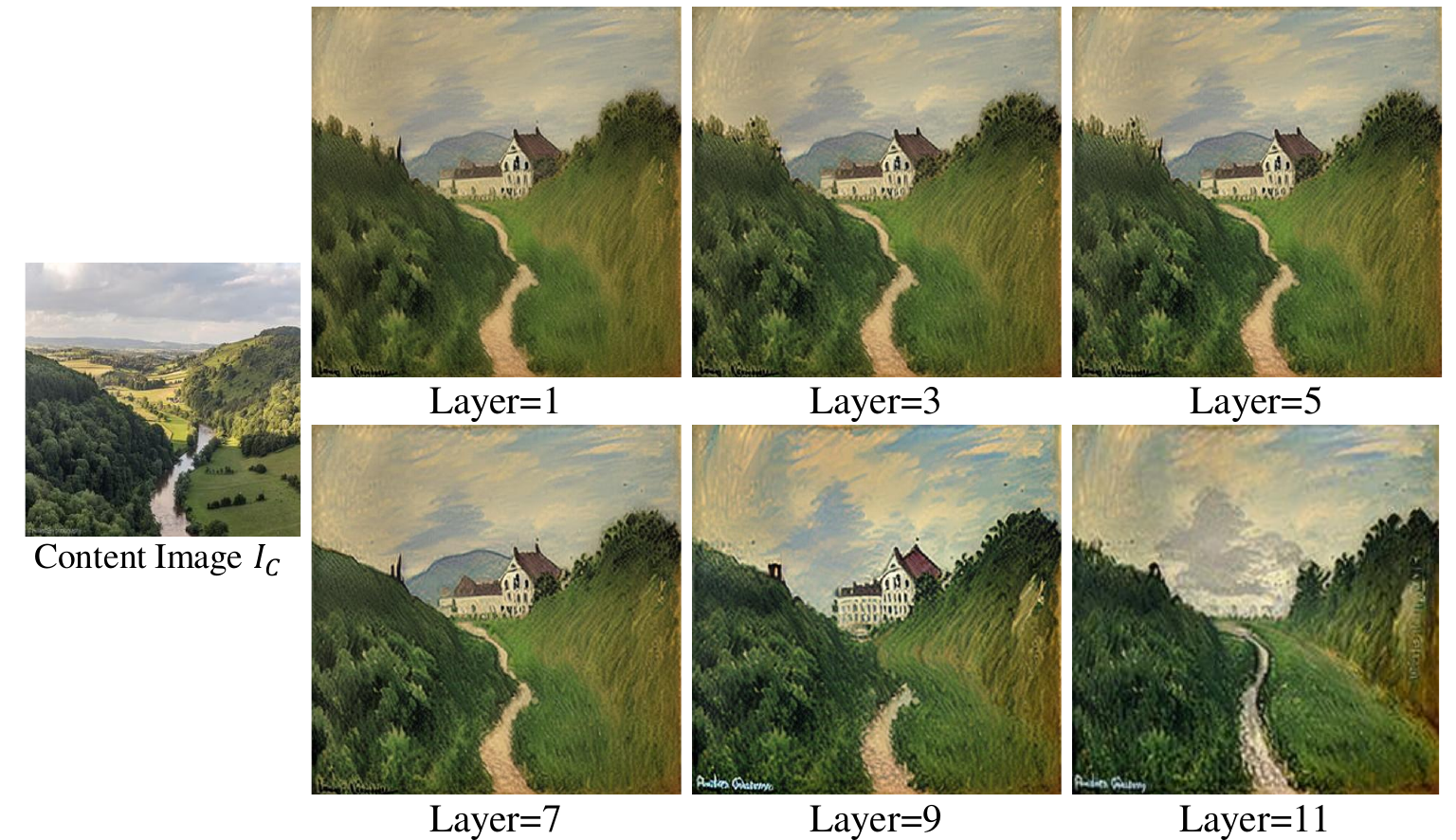}
\caption{Effect of layer level of CLIP encoder in Content Alignment Guidance for content preservation.}
\label{scale_trend}
\end{minipage}
\hfill
\begin{minipage}{0.48\linewidth}
\centering
\centering
\includegraphics[width=0.9\linewidth]{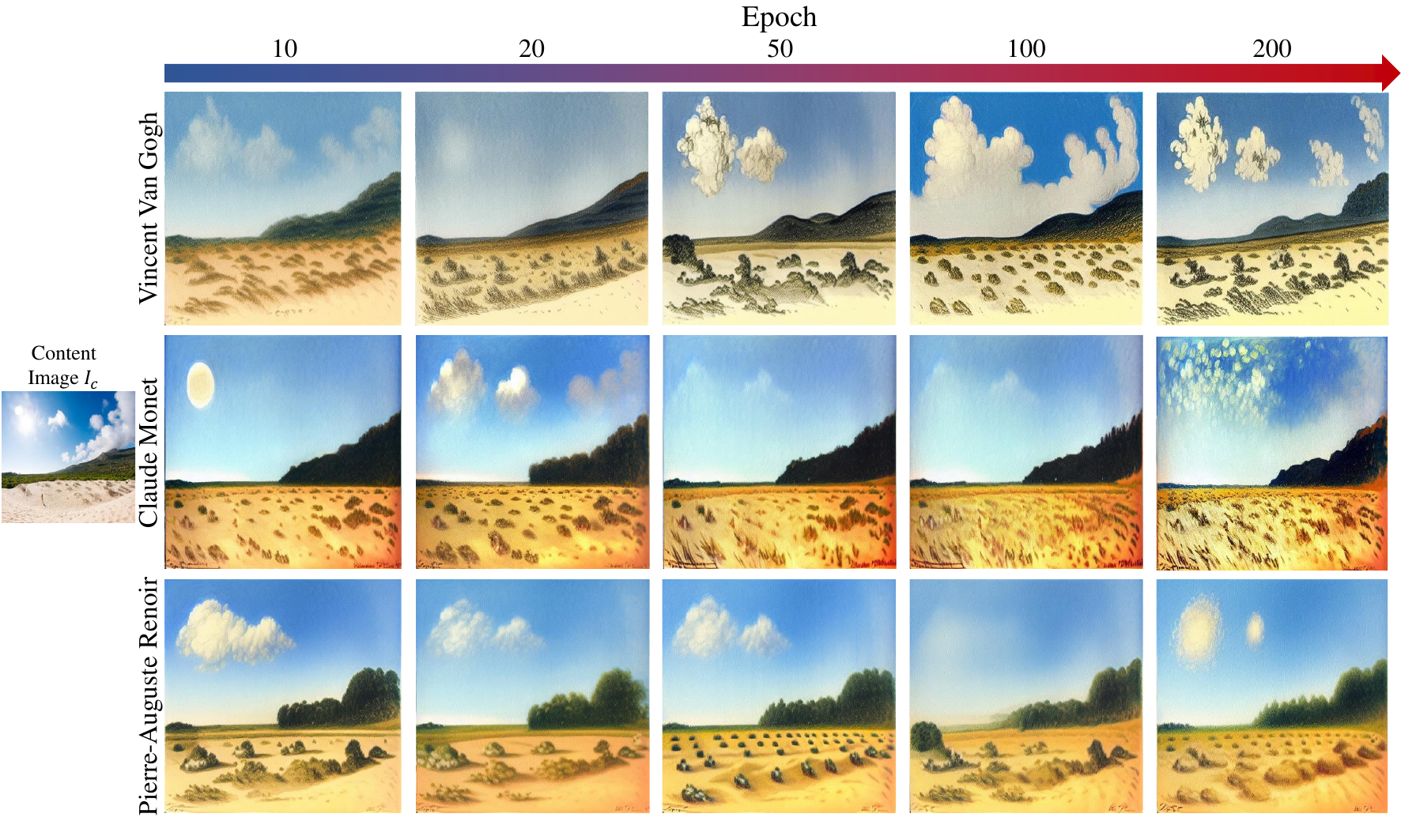}
\caption{Effect of training epochs on \emph{Style Extraction Function}: higher epochs yield stronger artist-specific characteristics.} 
\label{epoch}
\end{minipage}
\end{figure*}

To verify the effectiveness of CAG, we compare the full methods with a variant without CAG. As shown in \cref{fig:ablation,tab:ablation}, removing CAG leads to noticeable content degradation, reflected by a significant increase in ArtFID, which measures both style and content fidelity. These results demonstrate that CAG plays a crucial role in preserving content under strong global style modulation. Additional ablations on the embedding layer selection and training epochs are provided in \cref{scale_trend,epoch}.

\subsubsection{Effect of Layer Level for Content Alignment Guidance.}
We examine how applying Content Alignment Guidance (CAG) at different layers of the CLIP image encoder affects the preservation of content structure during global style transfer. As shown in \cref{scale_trend}, applying CAG at lower or mid-level layers (1–9) produces unintended artifacts, such as house-like structures absent from the content image $I_c$, indicating poor preservation of global composition. This occurs because lower layers mainly encode low-level features such as edges, color and textures. In contrast, applying CAG at Layer 11 preserves the original scene structure and yields semantically consistent results, showing that higher-level semantic features provide more reliable content alignment.

\subsubsection{Effect of Training Epochs on Style Extraction Function.}
We investigate the effect of training epochs on the \emph{Style Extraction Function} (SEF) $f_t$ to evaluate how training duration influences global style learning. We train SEF with five epoch settings (10, 20, 50, 100, 200) using artworks from three representative artists: Van Gogh, Monet, and Renoir. As shown in \cref{epoch}, models trained for only 10 or 20 epochs fail to capture distinctive stylistic characteristics, producing visually similar results across artists. As training progresses, the model gradually learns artist-specific brushwork, composition, and color distributions. At 200 epochs, the generated images exhibit clear stylistic separation, indicating that sufficient training is essential for learning a stable global style representation.



\section{Conclusion}
\label{sec:conclusion}
We propose \emph{Global Style Transfer}, a novel artistic image synthesis paradigm that represents the global artistic style of a target artist from multiple artworks. Through \emph{Global Style Guidance} and \emph{Content Alignment Guidance}, our method captures artist-level stylistic semantics while preserving the content structure and allowing style-driven geometric deformation. By moving beyond instance-level style transfer and text-dependent artistic synthesis, GST provides a more faithful and unbiased way to model an artist’s coherent visual identity.

\section*{Acknowledgment}
This work was partly supported by Institute of Information \& Communications Technology Planning \& Evaluation (IITP) grant funded by the Korea government (RS-2026-25516375), the Korea Institute of Science and Technology (KIST) Institutional Program (No. 26E0212). This work was supported by the IITP(Institute of Information \& Communications Technology Planning \& Evaluation)-ITRC(Information Technology Research Center) grant funded by the Korea government(Ministry of Science and ICT)(IITP-2026-RS-2023-00258649,33\%).


%
%
\clearpage
\bibliographystyle{splncs04}
\bibliography{main}

\clearpage

\appendix
\section{Experiment Settings}
\subsubsection{Model.}
We use Stable Diffusion v1.4 \cite{rombach2022high} as the default base generator for our Global Style Transfer experiments unless otherwise specified. To implement our proposed \emph{Content Alignment Guidance} (CAG), we use CLIP \cite{radford2021learning} as the image encoder, specifically the ViT-L/14 model, to extract high-fidelity content features.

\subsubsection{Hyper Parameter Settings.}
To produce our proposed \emph{Global Style Guidance} (GSG) on Text-to-Image latent diffusion model, we train the \emph{Style Extraction Function} (SEF) $f_t$ on the WikiArt dataset \cite{tan2018improved}, using between 500 and 1800 artworks per artist. Our proposed Style Extraction Function is implemented as an MLP with a single hidden layer (1280×1280). Training is conducted with a batch size of 8 during 200 training epochs. We use a learning rate of 0.1 with an Adam optimizer with $(\beta_1,\beta_2)$=(0.9, 0.999), weight decay is 0, and Adam epsilon is 1e\text{-}8. The impact of training epochs is presented in Fig. 7 of the main paper. Furthermore, we control the magnitude of the global style offset $\Delta \textbf{h}_t$ using a scaling factor $w$ in Eq. (5), for which we utilize values \{1.0, 1.25, 1.5\}, and we set the CAG guidance scale to $s$= 50.0. \cref{scale_trend} illustrates the qualitative effect of jointly varying both scales over a broader range ($w$ from 0.1 to 2.0 and $s$ from 1 to 80); within this range, our main experiments use $w \in \{1.0, 1.25, 1.5\}$ and $s$= 50. We recommend that selecting an appropriate scale is essential, as it reflects the trade-off between the proposed Global Style Guidance and Content Alignment Guidance.

\subsubsection{Computation Overhead.}
On a single NVIDIA RTX 3090 GPU, training the Style Extraction Function for one epoch takes approximately 216 seconds. The total training time scales with the number of artworks used per artist. After training, generating a 512$\times$512 image on the same GPU takes roughly 20 seconds.




\subsubsection{Evaluation Metrics.} 
To comprehensively evaluate the quality of our synthesized images, we employ five complementary metrics: FID, ArtFID, CFSD, CLIP-Div, and 1-Precision, each capturing a different aspect of style and content fidelity.

\noindent First, we measure Frechet Inception Distance (FID) \cite{heusel2017gans} between the generated images and the full set of artworks for each artist. This allows us to assess stylistic bias by quantifying how closely the synthesized samples align with the overall style distribution of the original artworks. Second, we report ArtFID \cite{wright2022artfid}, a metric designed to evaluate style-transfer performance by jointly considering content and style preservation. ArtFID reflects both resemblance to the artwork distribution and consistency with the content-image distribution, and is known to strongly correlate with human perceptual judgment. Following prior work, ArtFID is computed as $ArtFID=(1+LPIPS)\cdot(1+FID)$. Third, to isolate content fidelity from stylistic influence, we compute the Content Feature Structural Distance (CFSD) \cite{chung2024style}. CFSD measures structural similarity by capturing only the spatial correlations between local image patches, independent of stylistic appearance. Fourth, we measure CLIP-Diversity (CLIP-Div) to assess the stylistic spread of the generated images. Following ~\cite{alanov2023styledomain}, it is computed as the mean pairwise cosine distance between the CLIP image embeddings (ViT-L/14) of a generated set; higher values indicate greater stylistic diversity. In Tab. 2 of the main paper, we report it as (Ours/Real) to enable direct comparison with the real artwork corpus. Finally, we report 1-Precision to quantify memorization avoidance. Precision is the improved precision metric ~\cite{kynkaanniemi2019improved}, measuring the fraction of generated samples that fall within the real-artwork manifold, computed with the prdc implementation (k=5) ~\cite{naeem2020reliable}. A higher 1-Precision indicates less memorization of the training corpus.

\section{Further Analysis}
\begin{figure}[t!]
\centering
    \includegraphics[width=1.0\linewidth]{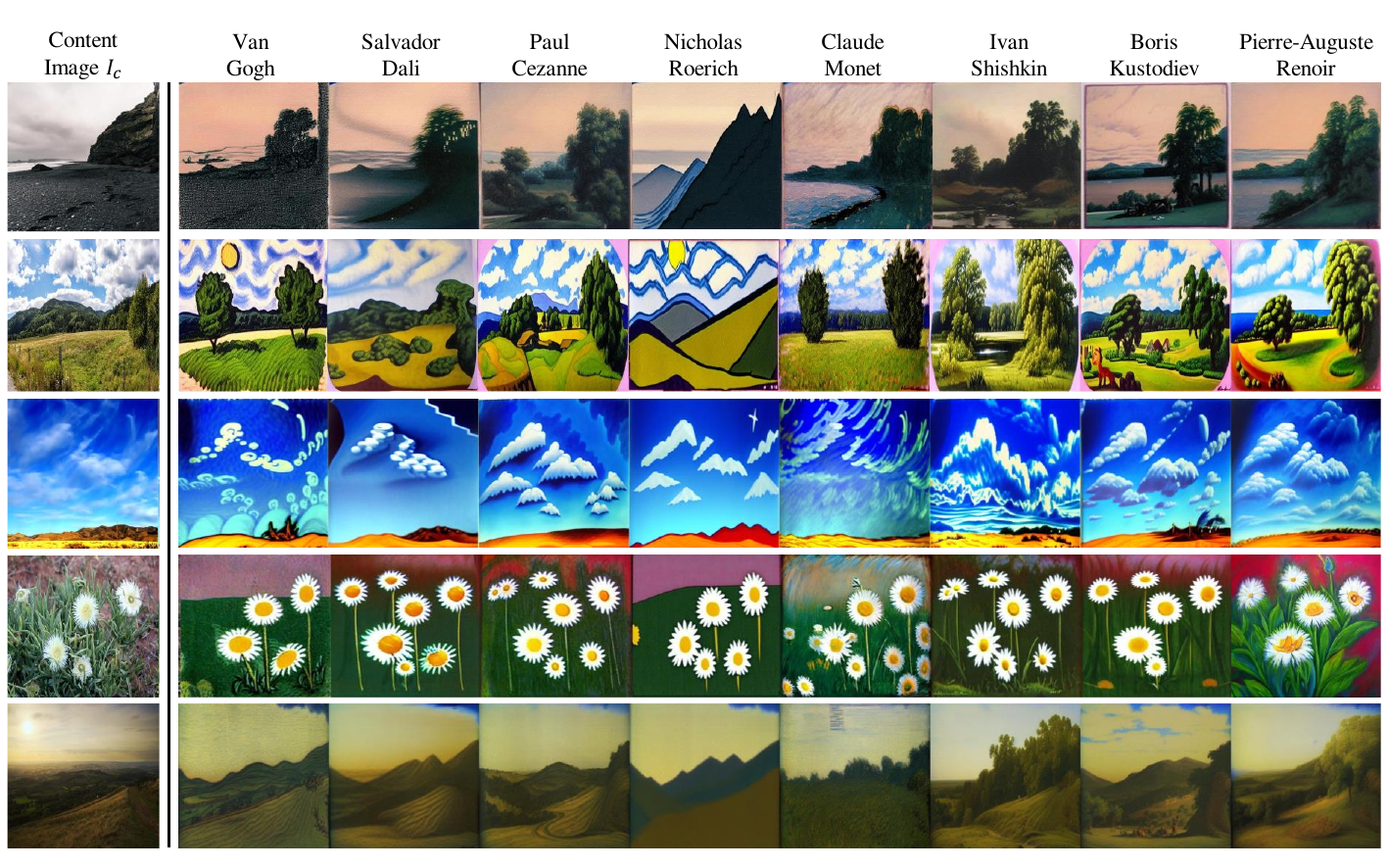}
    \caption{Global Style Transfer results across multiple artists for the same content image. Given the same content images $I_c$, our framework transfers the images into the global styles of eight different painters. All images are generated using a text-independent prompt (``\texttt{A painting}”) to eliminate stylistic bias in the diffusion model. For each content image $I_c$ (left column), our framework applies Global Style Guidance (GSG) to capture artist-specific global semantics and Content Alignment Guidance (CAG) to preserve flexible, style-based deformation of content. The results show that a single content image is rendered into distinctly different artistic styles across various artists, demonstrating our framework’s ability to synthesize artist-faithful images.}
\label{graphic_ab}
\end{figure}
\subsubsection{Global Style Transfer Results of Multiple Artists for the Same Content Image.}
We extend Fig. 1 of the main paper by visualizing stylization results across multiple artists for the same content image. \cref{graphic_ab} illustrates the results of applying our Global Style Transfer framework to the same content image $I_c$, where the image is stylized using the global styles of eight different painters.
For each artist, we generate 1,500 artistic images by applying the corresponding global style to the same set of content images. \cref{graphic_ab} shows representative examples from these results, where a randomly selected content image is stylized according to each artist’s global style. Despite sharing the same content image, the generated artistic images clearly reflect the distinctive global stylistic characteristics of each artist.



\begin{figure*}[t!]
\centering
\includegraphics[width=1.0\linewidth]{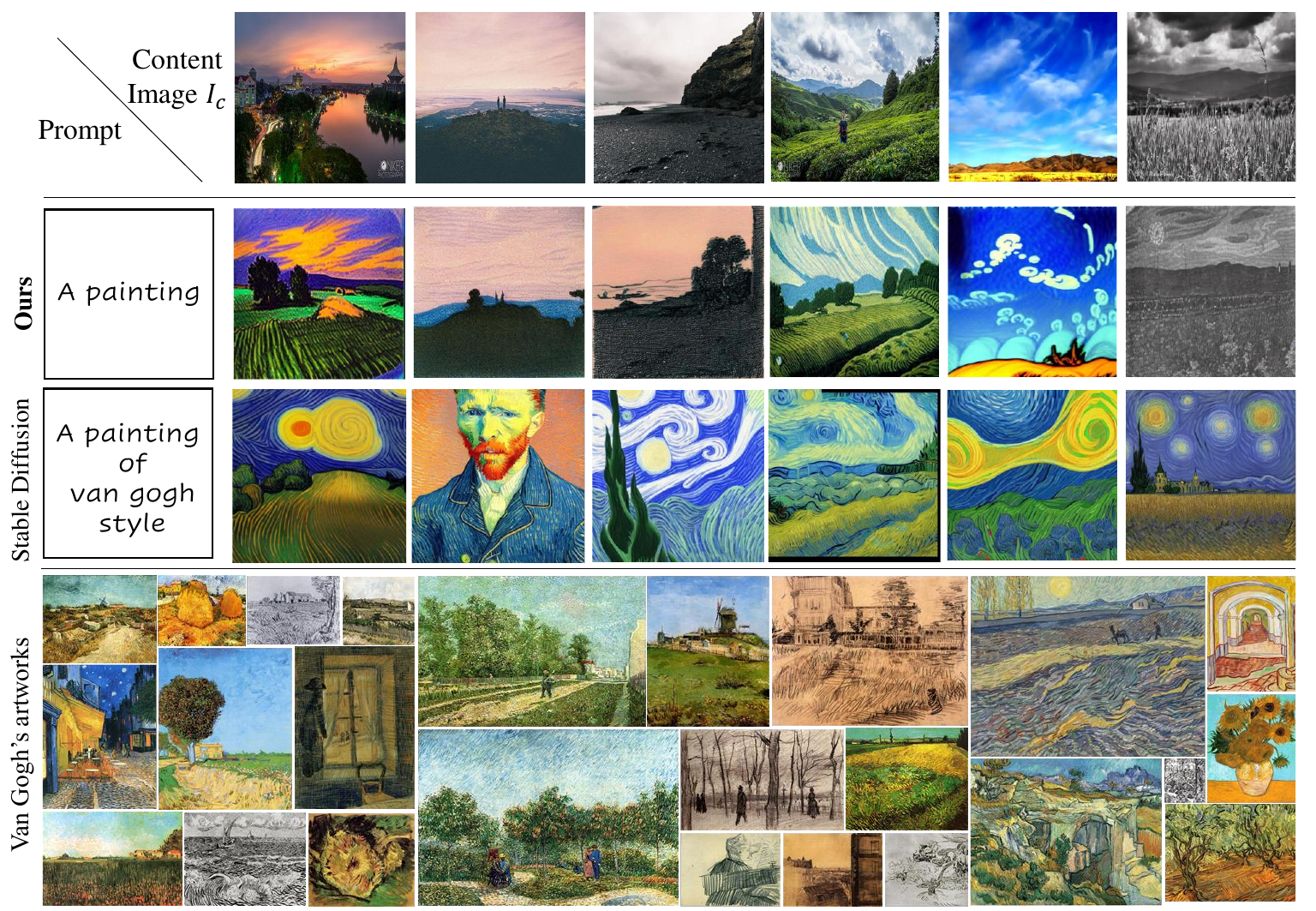}
\caption{Comparison of stylistic bias between vanilla Stable Diffusion (SD) and our Global Style Transfer. Given the same content images, vanilla SD conditioned on the prompt `\texttt{A painting of van gogh style'} tends to bias toward a few iconic patterns (e.g., `\emph{Starry Night}'-like swirls), often distorting the original structure and semantic content. In contrast, our method generates diverse stylizations while preserving the underlying content structure by learning a global stylistic representation from the full corpus of Van Gogh’s artworks. The bottom panel shows the real Van Gogh artwork distribution.
}
\label{vanilla}
\end{figure*}

\subsubsection{Further Analysis of Stylistic Bias on Stable Diffusion.}
Extending the experimental results of Fig. 7 of the main paper, we further examine whether stylistic bias also appears in vanilla Stable Diffusion \cite{rombach2022high}. Since our Global Style Transfer framework is built upon Stable Diffusion, we further compare our method with vanilla Stable Diffusion to investigate whether artistic stylistic bias also emerges in the original diffusion model. To conduct this analysis, we generate artistic images using both vanilla Stable Diffusion and our proposed Global Style Transfer framework. We use 1,500 photographs from the Vangogh2photo dataset \cite{zhu2017unpaired} as content images $I_c$ and obtain their latent representations via DDIM inversion. Our method produces stylized outputs from these latents using the prompt `\texttt{A painting}', whereas vanilla Stable Diffusion generates images conditioned on the prompt `\texttt{A painting of \{artist\} style}'.

Our analysis reveals that vanilla Stable Diffusion frequently collapses toward a small subset of iconic artworks. Furthermore, in many cases, the generated images lose the original content structure and semantic information, while being transformed into compositions that resemble specific well-known paintings by the artist. This phenomenon indicates that the model overemphasizes a few dominant stylistic patterns rather than capturing the broader stylistic distribution of the artist. To quantitatively evaluate this effect, we compute ArtFID between the generated images and the full artwork corpus for each artist, as reported in Tab. 2 in the main paper. Our method consistently achieves lower ArtFID for most artists, demonstrating improved alignment with the global artistic distribution. To further illustrate this phenomenon, \cref{vanilla} visualizes representative samples from the Van Gogh experiments, where the stylistic bias of vanilla Stable Diffusion becomes particularly evident. These results confirm that our method effectively mitigates stylistic bias and better preserves stylistic diversity across the artist’s corpus.

\begin{figure}[t!]
\centering
\includegraphics[width=0.6\linewidth]{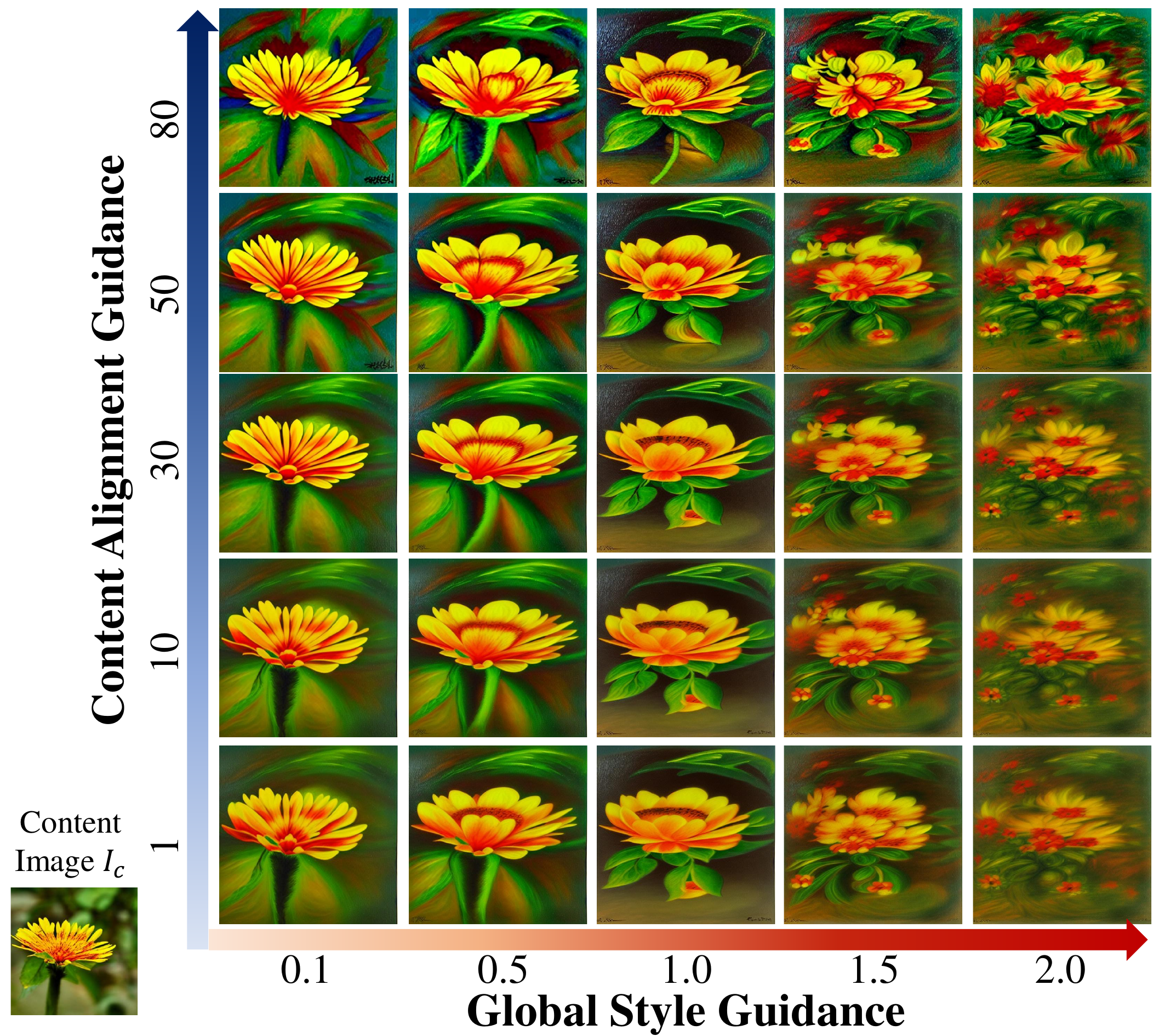}
\caption{Effect of guidance strength on the style–content balance, illustrating how Global Style and Content Alignment Guidance jointly control stylistic expressiveness and content preservation.}
\label{scale_trend}
\end{figure}
\begin{figure}[t]
\centering
  \includegraphics[width=0.5\linewidth]{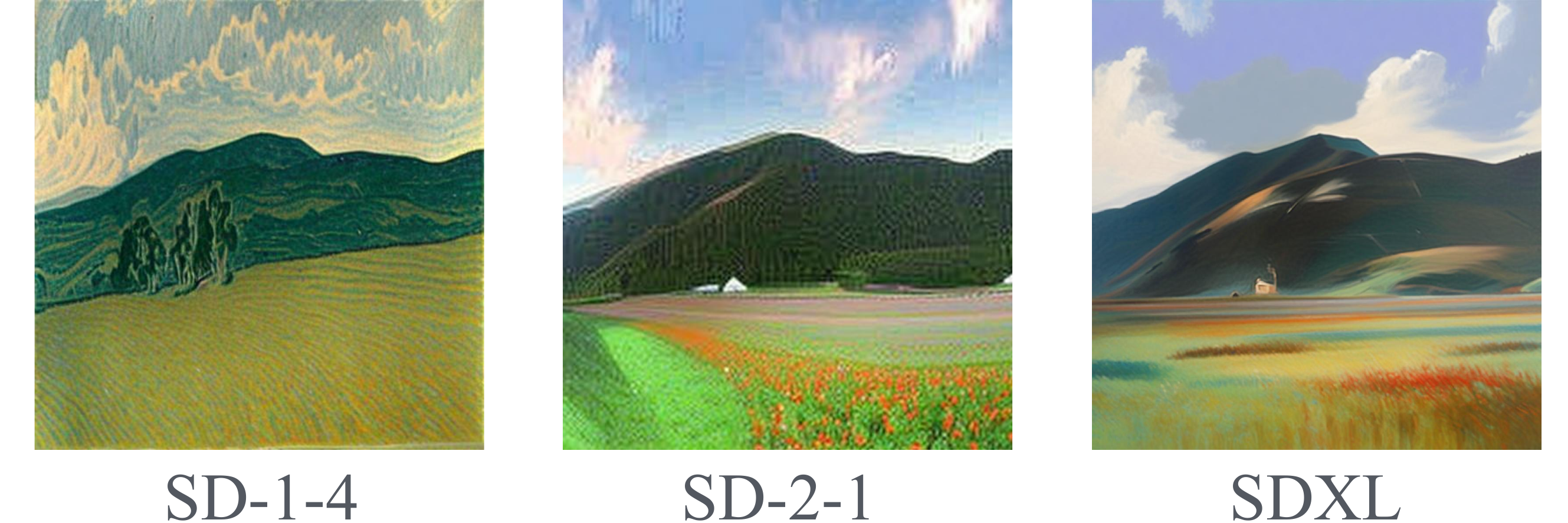}
  \caption{Backbone generalization. Global Style Transfer applied on different T2I diffusion backbones (SD-1.4, SD-2.1, SDXL) for the same content image, showing that our framework transfers consistently across backbones.}
  \label{fig:backbone}
\end{figure}
\subsubsection{Effect of Guidance Scales on Style-Content Trade Off.}
We qualitatively analyze the interplay between the \textit{Global Style Guidance} (GSG) and the \textit{Content Alignment Guidance} (CAG) in our framework. As shown in \cref{scale_trend}, increasing the \emph{Global Style Guidance} enhances stylistic expressiveness, yielding richer color saturation and brushstroke abstraction, while excessive strength may distort the geometric structure of the content. Conversely, stronger Content Alignment Guidance preserves the semantic integrity and spatial layout of the content but limits stylistic diversity. A balanced configuration of both guidance scales produces the most harmonious results, maintaining recognizable content structure while expressing vivid stylistic characteristics. This trade-off highlights that proper coordination between style and content guidance is essential for generating perceptually coherent and semantically consistent images.

\subsubsection{Sample Visualization with Different T2I Diffusion Backbones.}
To verify that Global Style Transfer is not tied to a specific base generator, we apply our framework on three different text-to-image diffusion backbones: Stable Diffusion v1.4, Stable Diffusion v2.1, and SDXL. Since the U-Net bottleneck (h-space) dimensionality differs across these backbones, we train a separate Style Extraction Function on the corresponding h-space for each backbone. As shown in~\cref{fig:backbone}, given the same content image, GST produces coherent artist-level stylization across all three backbones, indicating that our framework is applicable across different diffusion architectures rather than being tied to a specific one.

\section{Limitation}
\begin{figure}[t!]
\centering
\includegraphics[width=1.0\linewidth]{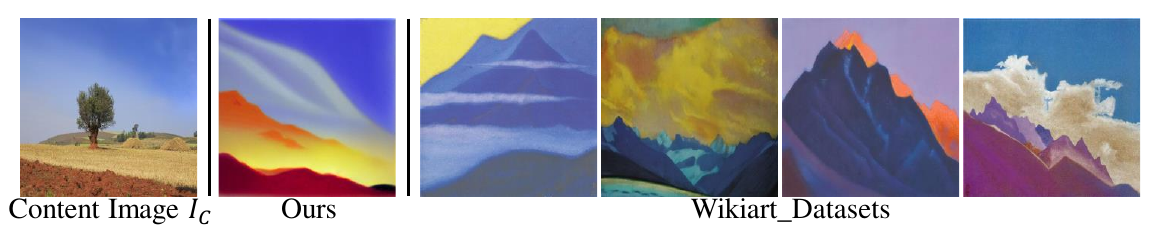}
\caption{Failure case of Global Style Transfer when the artist’s dataset contains highly biased content. As shown in Nicholas Roerich, whose works predominantly depict mountains, our framework fails to generalize to non-mountain content. When a tree-field landscape is provided as input, the model still reconstructs mountain-like shapes, reflecting the strong content bias encoded in the artist’s dataset.}
\label{failure}
\end{figure}

Our method relies on the visual diversity of an artist’s training data. When an artist’s works mostly depict limited subjects, such as natural scenes, the model struggles to generalize to unseen content like humans or vehicles, as shown in \cref{failure}. Thus, the effectiveness of style transfer depends on whether the artist’s dataset includes the target content type.
\section{Broader Impact}
Our Global Style Transfer framework enables artist-level style synthesis from large artwork collections, offering new possibilities for artistic creation, digital heritage preservation, and accessible content generation. It can support cultural institutions, educators, and creators by providing faithful stylistic reinterpretations without extensive generative AI model training. However, reproducing artistic styles at scale also raises concerns regarding authorship, cultural integrity, and responsible use of copyrighted material. We encourage the deployment of this technology in ways that respect artistic ownership and contribute positively to creative and cultural domains.

\end{document}